\pdfoutput=1
\documentclass[conference]{IEEEtran}
\IEEEoverridecommandlockouts
\usepackage{cite}
\usepackage{amsmath,amssymb,amsfonts}
\usepackage{algorithmic}
\usepackage{graphicx}
\usepackage{textcomp}
\usepackage{xcolor}
\usepackage{booktabs}
\usepackage{multirow}

\usepackage{amssymb} 
\usepackage{pifont}  

\def\BibTeX{{\rm B\kern-.05em{\sc i\kern-.025em b}\kern-.08em
    T\kern-.1667em\lower.7ex\hbox{E}\kern-.125emX}}
\begin{document}

\title{PoseCompass: Intelligent Synthetic Pose Selection for Visual Localization}

\author{
\IEEEauthorblockN{
Yanan Zhou\textsuperscript{1}, 
Zhaoyan Qian\textsuperscript{1}, 
Yanli Li\textsuperscript{1}, 
Nan Yang\textsuperscript{1}, 
Zhongliang Guo\textsuperscript{2}, 
Dong Yuan\textsuperscript{1*}
}

\IEEEauthorblockA{
\textsuperscript{1}The University of Sydney, Australia \\
\textsuperscript{2}University of St Andrews, United Kingdom \\
\{yzho5556, zqia6975\}@uni.sydney.edu.au, 
dong.yuan@sydney.edu.au
}
\thanks{Yanan Zhou and Zhaoyan Qian contributed equally }
\thanks{*Corresponding Author}
}

\maketitle

\begin{abstract}
In visual localization, Absolute Pose Regression (APR) enables real-time 6-DoF camera pose inference from single images, yet critically depends on fine-tuning data quality and coverage. While recent methods leverage 3D Gaussian Splatting (3DGS) for novel view synthesis–based data augmentation, random sampling generates redundant views and noisy samples from poorly reconstructed regions. To mitigate the research gap, we propose \textbf{PoseCompass}, an intelligent pose selection pipeline for 3DGS‑based APR. 
PoseCompass starts from formulating synthetic pose selection and deriving a value‑based pose ranking mechanism to identify informative poses. 
The ranking integrates three dimensions: (1) \textit{Localization Difficulty}, favoring challenging regions; (2) \textit{Coverage Novelty}, exploring under‑sampled areas; and (3) \textit{Rendering Observability}, filtering artifacts and noise. 
Then, PoseCompass generates trajectory‑constrained candidates, selects the top‑$K$ ranked poses, and synthesizes views using 3DGS with lightweight diffusion‑based alignment. 
Finally, the pose regressor is fine‑tuned on mixed real and synthetic data.
We evaluate our PoseCompass on 7-Scenes, PoseCompass reduces adaptation time from 15.2 to 5.1 minutes ($3\times$ speedup) while cutting median pose errors by 53.8\%, significantly outperforming random baselines. 

\end{abstract}


\section{Introduction}
\label{sec:intro}

Visual localization is fundamental to embodied AI systems, including indoor navigation and autonomous driving. Absolute Pose Regression (APR)~\cite{kendall2015posenet, brahmbhatt2018geometry, chen2024mapose, Li2025unleashing} predicts 6-DoF camera poses via end-to-end neural networks, circumventing the high computational overhead of feature matching~\cite{sarlin2020superglue, lindenberger2023lightglue} and PnP solvers~\cite{gao2003complete}. However, APR models often learn implicit image-to-pose retrieval rather than genuine geometric understanding~\cite{Sattler_2019_CVPR}, leading to poor generalization, particularly under sparse training data regimes.

\begin{figure}[!t]
  \centering
  \includegraphics[width=0.7\columnwidth]{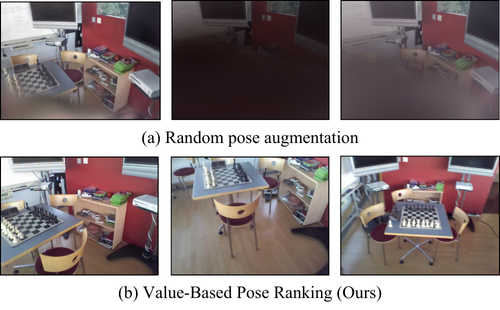}
  \caption{\textbf{Value-Based Pose Selection.} 
  (a) Random sampling produces redundant and low-quality views. 
  (b) Our value-based ranking selects diverse, high-quality poses.}
  \label{fig:teaser}
\end{figure}

Novel View Synthesis (NVS) provides an effective mechanism for augmenting training data in absolute pose regression. While NeRF-based approaches~\cite{moreau2022lens, chen2022dfnet, mildenhall2020nerf} can expand viewpoint coverage, their high rendering cost limits scalability in practical adaptation scenarios. Recently, 3D Gaussian Splatting (3DGS)~\cite{3dgs} has emerged as a more efficient alternative, enabling real-time novel view rendering. Despite this advantage, existing methods largely leave pose selection under-specified. In practice, synthetic poses are often sampled randomly, which leads to two systematic issues: redundant viewpoints that contribute limited additional supervision, and noisy samples arising from poorly reconstructed regions like texture-less surfaces or scene boundaries. These effects reduce data efficiency under constrained rendering budgets, shown as Fig.~\ref{fig:teaser}(a).

Data augmentation via NVS has shown promise for APR training. Recent work~\cite{Li2025unleashing} leverages 3DGS for data augmentation and restricts sampling to trajectory-constrained neighborhoods around training poses to avoid severe extrapolation. However, within these constrained regions, pose selection remains random, and the issues of redundancy and rendering-induced noise are not explicitly addressed.

In this paper, we introduce \textbf{PoseCompass}, an intelligent pipeline for synthetic pose selection in 3DGS‑based absolute pose regression. 
Inspired by gradient‑based active learning~\cite{ash2020deep}, we formulate pose selection as a budget‑constrained optimization problem that aims to maximize expected error reduction under limited rendering budgets. 
To this end, we derive a value‑based pose ranking function that integrates three complementary dimensions: 
(1) \emph{Localization Difficulty}, which captures model weaknesses through zero‑shot prediction errors to prioritize challenging regions; 
(2) \emph{Coverage Novelty}, which measures spatial distances to existing samples to encourage diverse viewpoint coverage; and 
(3) \emph{Rendering Observability}, which evaluates 3DGS reconstruction quality via geometric consistency to suppress artifacts and unreliable renderings~\cite{brachmann2023accelerated}. 
Based on this value‑based ranking, our pipeline generates trajectory‑constrained pose candidates, selects informative poses for 3DGS‑based view synthesis with Syn2Real alignment, and fine‑tunes the pose regressor on mixed real and synthetic data. 
The resulting synthetic views are illustrated in Fig.~\ref{fig:teaser}(b). 
Extensive experiments on indoor (7‑Scenes~\cite{shotton2013scene}) and outdoor (Cambridge Landmarks~\cite{kendall2015posenet}) benchmarks demonstrate that PoseCompass achieves state‑of‑the‑art localization accuracy, reducing median pose errors by over 50\% while delivering a $3\times$ training speedup over random pose augmentation.

In summary, our contributions are:
\begin{itemize}
    \item We formalize pose selection for 3DGS-based APR data augmentation as a budget-constrained optimization problem and derive a value function grounded in active learning to identify the most valuable synthetic training poses.
    \item We propose a value-based pose ranking mechanism with the multiplicative fusion of localization difficulty, coverage novelty, and rendering observability for intelligent subset selection.
    \item We design PoseCompass, an pipeline integrating value-based ranking, 3DGS rendering, Syn2Real alignment, and efficient fine-tuning, setting a new benchmark for data-efficient APR training.
\end{itemize}

\begin{figure*}[ht]
    \centering
    \includegraphics[width=0.92\textwidth]{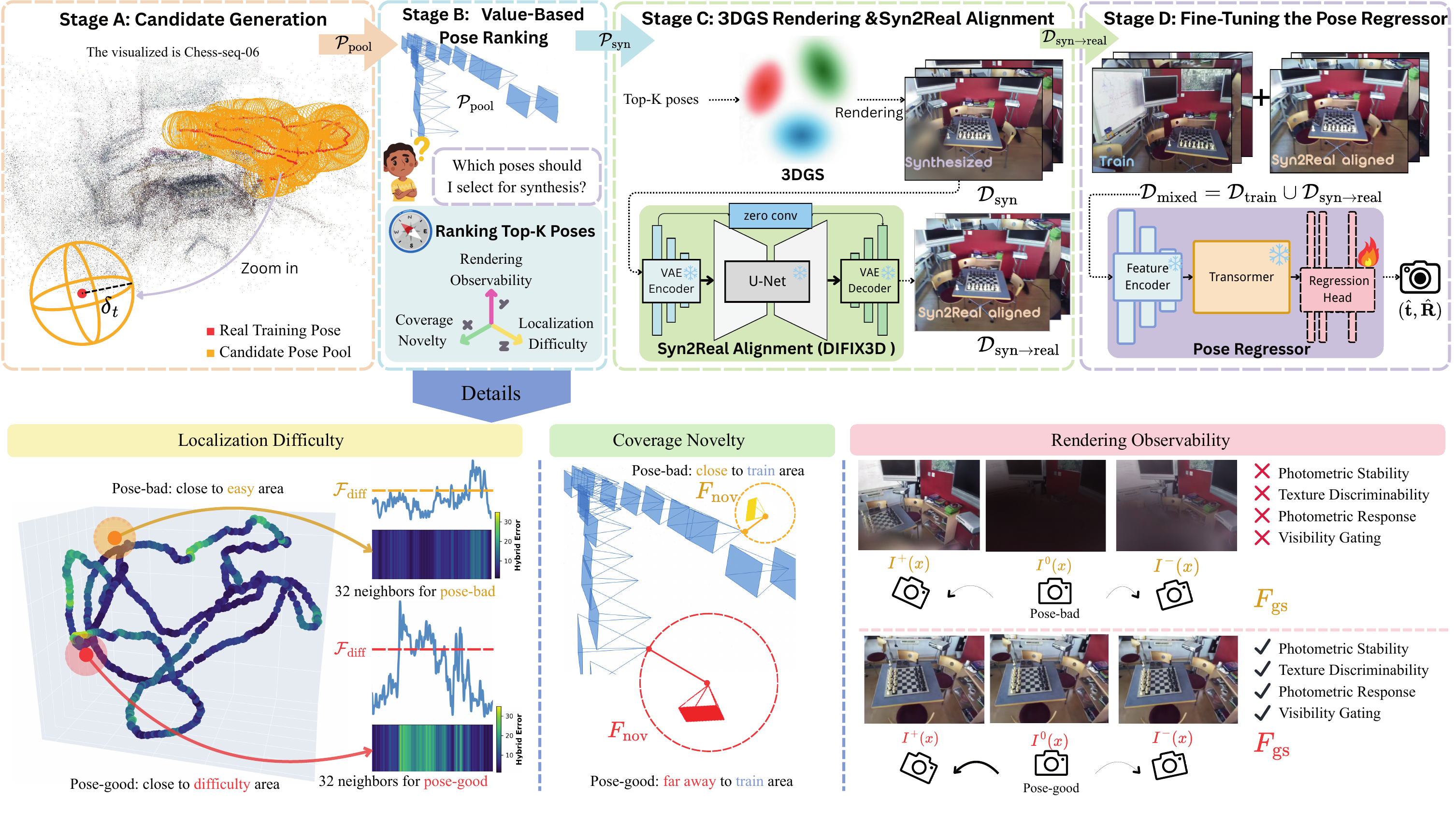}
    \caption{
        The proposed PoseCompass stages.
    }
    \label{fig:main}
\end{figure*}

\section{Related Works}
\label{sec:relatedworks}

\textbf{Visual localization.}
Visual localization aims to estimate a camera's translation and rotation in a 3D scene.
Classical geometric approaches~\cite{sarlin2020superglue, lindenberger2023lightglue}typically build point clouds and maintain a reference image database, then perform image retrieval with precomputed descriptors to establish 2D--3D correspondences.
A closely related line of work, scene coordinate regression (SCR)~\cite{brachmann2017dsac, brachmann2023accelerated, wang2024glace, brachmann2021visual}, uses neural networks to directly predict 2D--3D mappings.
However, both families require solving PnP at inference time~\cite{gao2003complete}, which substantially increases computation.
Absolute pose regression (APR)~\cite{kendall2015posenet, brahmbhatt2018geometry, Kendall_Cipolla_2017, Moreau, Chen_Wang_Prisacariu_2021, Li2025unleashing}
directly regresses poses from images in an end-to-end manner, enabling efficient inference, but still struggles to generalize across scenes due to long training and limited transfer.
Within the APR paradigm, the latest map-relative pose regression method~\cite{chen2024mapose} pre-trains a pose regressor on hundreds of scenes and performs minute-level scene-specific adaptation, markedly improving deployability.

\textbf{Data augmentation for pose regression.}
The effectiveness of APR highly depends on the scale and diversity of training data.
Prior work~\cite{Sattler_2019_CVPR} shows that APR models often learn an implicit ``image-to-pose retrieval'' relationship rather than genuine scene understanding.
Motivated by this, many methods use novel view synthesis (NVS) to expand training viewpoints and increase correspondence coverage.
Subsequent approaches, including LENS~\cite{moreau2022lens}, DFNet~\cite{chen2022dfnet}, and PMNet~\cite{lin2024learning}, augment data via NeRF, but NeRF's training and rendering inefficiency hinders deployment.
With its efficient scene modeling and controllable appearance rendering, 3DGS~\cite{3dgs} has emerged as a stronger alternative to NeRF.
RAP~\cite{Li2025unleashing} leverages 3DGS with controllable appearance to notably improve the robustness of synthesized views under appearance changes, yet it still lacks clear principles for selecting which viewpoints to synthesize.
To select high valued synthesized view poses in 3DGS, our framework performs data-driven pose value estimation and sampling, enabling more efficient and robust augmentation for APR.

\section{Method}
\label{sec:method}


We formulate synthetic data augmentation for APR as selecting an optimal subset $\mathcal{S}^* \subseteq \mathcal{P}_{\text{pool}}$ of $K$ poses that maximizes training improvement:
\begin{equation}
\label{eq:optimization_objective}
\mathcal{S}^* = \arg\max_{|\mathcal{S}|=K} \sum_{P \in \mathcal{S}} \text{Value}(P),
\end{equation}
where $\text{Value}(P)$ measures the expected error reduction from adding $P$ to training set $\mathcal{D}_{\text{real}} = \{(\mathbf{I}_i, \mathbf{P}_i)\}_{i=1}^{N}$. Following gradient-based active learning~\cite{ash2020deep}, this decomposes as $\text{Value}(P) \propto I(P) \times D(P) \times Q(P)$, where $I(P)$, $D(P)$, $Q(P)$ represent information gain, diversity, and quality gate. We approximate these via efficient proxies---\emph{Localization Difficulty} ($\mathcal{F}_{\text{diff}}$), \emph{Coverage Novelty} ($\mathcal{F}_{\text{nov}}$), and \emph{Rendering Observability} ($\mathcal{F}_{\text{gs}}$)---yielding:
\begin{equation}
\label{eq:value_function}
V(\mathbf{P}) = \widetilde{\mathcal{F}}_{\text{diff}}(\mathbf{P})^\alpha \times (1+\widetilde{\mathcal{F}}_{\text{nov}}(\mathbf{P}))^\beta \times \widetilde{\mathcal{F}}_{\text{gs}}(\mathbf{P})^\gamma,
\end{equation}
where $(\alpha, \beta, \gamma)$ are weighting exponents. This enables greedy top-$K$ selection by ranking all candidates via $V(\cdot)$.

We design the PoseCompass (Fig.~\ref{fig:main}) as a four-stage pipeline:
(1) Trajectory-Constrained Candidate Generation (§\ref{sec:candidate_generation});
(2) Value-Based Pose Ranking (§\ref{sec:value_assessment});
(3) 3DGS Rendering and Syn2Real Alignment (§\ref{sec:selection_alignment});
(4) Fine-Tuning the Pose Regressor (§\ref{sec:fine_tuning}).

\subsection{Trajectory-Constrained Candidate Generation}
\label{sec:candidate_generation}

As shown in Stage 1 of Fig.~\ref{fig:main}, the training set contains $N$ real poses (red markers), which serve as trajectory constraints to delimit the scope of the synthetic pose pool. Following~\cite{Li2025unleashing}, for each real pose $\mathbf{P}_i \in \mathcal{P}_{\text{train}}$, we generate $M$ synthetic poses by adding uniform noise: translation $\delta_t \in [-0.20, 0.20]$m and rotation $\delta_r \in [-10^\circ, 10^\circ]$ for indoor (outdoor: $\delta_t{=}1.50$m, $\delta_r{=}4^\circ$). This yields a candidate pool $\mathcal{P}_{\text{pool}}$ (orange markers) with $M \times N \approx 10K$ candidates. The zoom-in view visualizes this process on Chess-seq-06, where candidates densely surround real training poses.

\subsection{Value-Based Pose Ranking}
\label{sec:value_assessment}

Stage 2 of Fig.~\ref{fig:main} illustrates our value-based ranking mechanism. Random sampling from $\mathcal{P}_{\text{pool}}$ suffers from redundant poses and low-quality renderings. We evaluate candidates via a value function integrating three complementary dimensions (detailed in bottom panels of Fig.~\ref{fig:main}): localization difficulty, coverage novelty, and rendering observability.

\subsubsection{Localization Difficulty}

The bottom-left panel of Fig.~\ref{fig:main} visualizes this dimension. We prioritize poses with high localization difficulty using zero-shot prediction errors. The heatmap compares pose-good (near difficulty areas, exhibiting high $\mathcal{F}_{\text{diff}}$ from 32 neighbors) with pose-bad (near easy areas with low scores). Following~\cite{Sattler_2019_CVPR}, we use the hybrid distance:
\begin{equation}
\label{eq:hybrid_metric}
d(\mathbf{P}_i, \mathbf{P}_j) = \frac{\|\mathbf{t}_i - \mathbf{t}_j\|_2}{\sigma_t} + \lambda \,\vartheta(\mathbf{R}_i, \mathbf{R}_j),
\end{equation}
where $\sigma_t$ is the translation standard deviation, $\vartheta(\mathbf{R}_i,\mathbf{R}_j){=}\arccos\bigl(\tfrac{\text{tr}(\mathbf{R}_i^\top\mathbf{R}_j){-}1}{2}\bigr)$ measures rotation distance, and $\lambda{=}1.0$ balances translation and rotation components. For candidate $\mathbf{P}$, we aggregate zero-shot errors from $k{=}32$ nearest training poses via Gaussian kernel weighting, then apply quantile normalization to obtain $\widetilde{\mathcal{F}}_{\text{diff}} \in [0,1]$.

We emphasize that localization difficulty is computed in a zero-shot manner and used solely for offline pose ranking, rather than as a training signal. Fine-tuning on mixed real and synthetic data enables the regressor to correct initial biases, and the consistent gains across scenes indicate that the criterion captures transferable failure modes rather than model-specific artifacts.

\subsubsection{Coverage Novelty}

The bottom-middle panel of Fig.~\ref{fig:main} demonstrates the novelty dimension. Pose-good (far from training area) receives higher $\mathcal{F}_{\text{nov}}$ scores, while pose-bad (close to training area) scores lower. To encourage spatial diversity, we define novelty as the minimum distance to existing training poses: $\mathcal{F}_{\text{nov}}(\mathbf{P}) = \min_{j} d(\mathbf{P},\mathbf{P}_j)$ using Eq.~\eqref{eq:hybrid_metric}. Quantile normalization yields $\widetilde{\mathcal{F}}_{\text{nov}} \in [0,1]$.

\begin{table*}[t]
\centering
\small
\caption{\textbf{Pose Estimation Results on 7-Scenes Dataset.} We Report Median Translation/Rotation Error (cm/°). Best Results in \textbf{Bold}, Second Best \underline{Underlined}.}
\label{tab:7scenes_main}
\setlength{\tabcolsep}{4pt}
\begin{tabular}{l|c|ccccccc|c}
\toprule
\textbf{Methods} & \textbf{Source} & \textbf{Chess} & \textbf{Fire} & \textbf{Heads} & \textbf{Office} & \textbf{Pumpkin} & \textbf{Kitchen} & \textbf{Stairs} & \textbf{Average} \\
\midrule
PoseNet~\cite{kendall2015posenet} & ICCV15 & 32/8.12 & 47/14.4 & 29/1.20 & 48/7.68 & 47/8.42 & 59/8.64 & 47/13.80 & 44/10.4 \\
MapNet~\cite{brahmbhatt2018geometry} & CVPR18 & 8/3.25 & 27/11.7 & 18/13.3 & 17/5.15 & 22/4.02 & 23/4.93 & 30/12.1 & 21/7.77 \\
MS-Transformer~\cite{shavit2021learning} & ICCV21 & 11/4.66 & 24/9.60 & 14/12.2 & 17/5.66 & 18/4.44 & 17/5.94 & 17/5.94 & 18/7.28 \\
PAE~\cite{shavit2022camera} & ECCV22 & 12/4.95 & 24/9.31 & 14/12.5 & 19/5.79 & 18/4.89 & 18/6.19 & 25/8.74 & 19/7.48 \\
DFNet~\cite{chen2022dfnet} & ECCV22 & 5/1.88 & 17/6.45 & 6/3.63 & 8/2.48 & 10/2.78 & 22/5.45 & 16/3.29 & 12/3.71 \\
PMNet~\cite{lin2024learning} & ECCV24 & 4/1.70 & 10/4.51 & 7/4.23 & 7/1.96 & 14/3.33 & 14/3.36 & 16/3.62 & 10/3.24 \\
RAP~\cite{Li2025unleashing} & ICCV25 & \underline{2/0.85} & 6/3.45 & 5/5.49 & 5/1.94 & 4/1.70 & 7/2.12 & 9/2.11 & 5/2.52 \\
MaRepo ~\cite{chen2024mapose} & CVPR24 & 2.6/1.35 & 2.5/1.42 & 2.3/2.21 & \underline{3.6/1.44} & 4.2/1.55 & 5.1/1.99 & 6.7/1.83 & 3.9/1.68 \\
MaRepo$_s$ (Random Aug)~\cite{chen2024mapose} & CVPR24 & 2.1/1.24 & \underline{2.3/1.39} & \underline{1.8/2.03} & 3.8/1.26 & \underline{3.5/1.48} & \underline{4.2/1.71} & \underline{5.6/1.67} & \underline{3.2/1.54} \\
\midrule
\textbf{PoseCompass} & \textbf{Ours} & \textbf{1.2/0.65} & \textbf{1.4/0.81} & \textbf{1.1/1.15} & \textbf{1.8/0.72} & \textbf{1.9/0.88} & \textbf{2.3/1.05} & \textbf{3.1/0.98} & \textbf{1.8/0.89} \\
\bottomrule
\end{tabular}
\end{table*}

\subsubsection{Rendering Observability in 3DGS}

The bottom-right panel of Fig.~\ref{fig:main} visualizes our quality assessment. Pose-good passes all four self-supervised checks (photometric stability \checkmark, texture discriminability \checkmark, photometric response \checkmark, visibility gating \checkmark), while pose-bad exhibits rendering artifacts (all \ding{55}). The preceding scores operate solely in pose space, ignoring 3DGS rendering quality. We train original 3DGS~\cite{3dgs} on $\mathcal{D}_{\text{train}}$. Since candidates lack ground truth, we assess quality via self-supervised geometric consistency: well-reconstructed regions exhibit smooth photometric transitions under small pose perturbations, while artifacts cause abrupt changes.


For each candidate $\mathbf{P}$, we render a central view $\mathbf{I}^{0}$ and perturbed views $\mathbf{I}^{\pm}$ with $\pm$2cm translation and $\pm$2° rotation offsets. We quantify photometric stability via minimum SSIM:
\begin{equation}
s = \min\!\bigl(\text{SSIM}(\mathbf{I}^{+},\mathbf{I}^{0}),\ \text{SSIM}(\mathbf{I}^{-},\mathbf{I}^{0})\bigr),
\end{equation}
which vetoes poses where any perturbation reveals artifacts. We compute scene-normalized gradient magnitude at each pixel $x \in \Omega$ (where $\Omega$ denotes the image domain):
\begin{equation}
\hat{g}(x) = \frac{\|\nabla \mathbf{I}^{0}(x)\|}{\overline{g} + \varepsilon}, \quad \overline{g} = \frac{1}{|\Omega|}\sum_{y \in \Omega} \|\nabla \mathbf{I}^{0}(y)\|,
\end{equation}
where $\varepsilon{=}10^{-8}$ ensures numerical stability and normalization by mean gradient $\overline{g}$ enables cross-scene comparability, prioritizing textured regions. The symmetric photometric response:
\begin{equation}
r(x) = \frac{\bigl|\mathbf{I}^{+}(x)-\mathbf{I}^{-}(x)\bigr|}{\bigl|\mathbf{I}^{+}(x)\bigr|+\bigl|\mathbf{I}^{-}(x)\bigr|+\varepsilon}
\end{equation}
measures intensity variation rate under perturbations, where well-reconstructed regions exhibit moderate values reflecting geometric parallax while artifacts produce abnormally high or near-zero responses. 

To suppress unobserved pixels, the visibility mask $w(x) = \sigma_{\tau_\alpha}\!\bigl(\alpha(x)\bigr) \cdot \sigma_{\tau_b}\!\bigl(b(x)\bigr) \cdot \sigma_{\tau_h}\!\bigl(h(x)\bigr)$ soft-gates by 3DGS opacity $\alpha(x)$, brightness $b(x)$, and local entropy $h(x)$ via sigmoid thresholds $\sigma_{\tau}(z){=}(1{+}e^{-k(z{-}\tau)})^{-1}$ with $k{=}10$ and $(\tau_\alpha,\tau_b,\tau_h){=}(0.5,0.2,3.0)$. The observability score integrates these factors:
\begin{equation}
\mathcal{F}_{\text{gs}}(\mathbf{P}) = s \cdot \frac{1}{|\Omega|}\sum_{x\in\Omega} w(x)\,\hat{g}(x)\,r(x),
\end{equation}
where global stability $s$ vetoes unstable poses and the weighted sum evaluates discriminability-weighted photometric variation under visibility constraints. This suppresses three failure modes: (i) textureless/underexposed regions (low $w(x)$, $\hat{g}(x)$), (ii) artifacts causing inconsistent perturbations (high $r(x)$ but low $s$, product vanishes), and (iii) unobserved areas (low $w(x)$). Quantile normalization yields $\widetilde{\mathcal{F}}_{\text{gs}} \in [0,1]$ for fusion with other dimensions.

After computing the three normalized scores, we rank candidates by $V(\mathbf{P})$ (Eq.~\eqref{eq:value_function}) and select the top-$K$ poses as $\mathcal{P}_{\text{syn}}$.

\subsection{3DGS Rendering and Syn2Real Alignment}
\label{sec:selection_alignment}

Stage 3 of Fig.~\ref{fig:main} shows the rendering and alignment process. After selecting $\mathcal{P}_{\text{syn}}$, we render the views in 3DGS. To mitigate the appearance gap between renderings and real images, we apply Difix3D+~\cite{difix3d}, a single-step diffusion aligner that improves photometric consistency while preserving pose labels:
\begin{equation}
\mathcal{D}_{\text{syn}\rightarrow\text{real}} = G_{\theta}(\mathcal{D}_{\text{syn}}).
\end{equation}

\subsection{Fine-Tuning the Pose Regressor}
\label{sec:fine_tuning}

In Stage 4 (Fig.~\ref{fig:main}), we adopt MaRepo~\cite{chen2024mapose}, a ViT-based pose regressor pre-trained on hundreds of scenes. We freeze the backbone and fine-tune only the lightweight regression head on $\mathcal{D}_{\text{mixed}}=\mathcal{D}_{\text{train}} \cup \mathcal{D}_{\text{syn}\rightarrow\text{real}}$ using $L_1$ pose loss: $\mathcal{L}_{\text{pose}} = \mathbb{E}\bigl[\|\hat{\mathbf{R}} - \mathbf{R}\|_{1} + \|\hat{\mathbf{t}} - \mathbf{t}\|_{1}\bigr]$. The syn2real aligner is used only for offline data preparation; inference retains the frozen backbone with the adapted head, preserving deployment efficiency.

\section{Experiments}
\label{sec:experiments}

\subsection{Evaluation Setup}

\textbf{Datasets.} Following prior works~\cite{lin2024learning, chen2024mapose, Li2025unleashing}, we evaluate on 7-Scenes~\cite{shotton2013scene}, an indoor dataset with seven scenes (1--18 m³) featuring texture-less regions, repetitive patterns, lighting changes, and occlusions. We use the standard split with SfM poses as ground truth~\cite{brachmann2017dsac, kendall2015posenet}. We also test on four scenes from Cambridge Landmarks~\cite{kendall2015posenet} (about 875 m²), an outdoor dataset with long baselines, viewpoint/illumination variations, moving objects, and similar challenging conditions.

\textbf{Baseline Methods.} We compare the proposed PoseCompass method with state-of-the-art single-frame APR approaches. Among these, PMNet~\cite{lin2024learning} and RAP~\cite{Li2025unleashing} are the most relevant methods based on data augmentation, while MaRepo~\cite{chen2024mapose} represents the most powerful absolute pose regression model. We also compare against MaRepo$_s$ (Random Aug), which applies the scene-specific fine-tuning with random augmentation as described in the original work.

\textbf{Implementation Details.} Following §\ref{sec:candidate_generation}, we generate candidates with perturbations $\delta_t{=}20$cm, $\delta_r{=}10^\circ$ (indoor) and $\delta_t{=}150$cm, $\delta_r{=}4^\circ$ (outdoor), yielding $\mathcal{P}_{\text{pool}}$ with $M{\times}N{\approx}10K$ poses per scene. We train 3DGS~\cite{3dgs} on $\mathcal{D}_{\text{train}}$. Using the value function (Eq.~\eqref{eq:value_function}) with exponents $(\alpha,\beta,\gamma){=}(1.5,1.0,2.0)$, we rank and select top-$K$ poses. Rendered views are aligned via frozen Difix3D+~\cite{difix3d} (§\ref{sec:selection_alignment}), then we fine-tune MaRepo~\cite{chen2024mapose} for 5 epochs on $\mathcal{D}_{\text{mixed}}$ (§\ref{sec:fine_tuning}). See supplementary for full details.

\subsection{Benchmark Results}
\textbf{7-Scenes~\cite{shotton2013scene}.} 
Table~\ref{tab:7scenes_main} presents the quantitative results on the 7-Scenes dataset. PoseCompass achieves state-of-the-art performance across all seven scenes, reducing the average median error to 1.8~cm / 0.89$^{\circ}$. Notably, our method significantly outperforms the previous SOTA method RAP~\cite{Li2025unleashing} (5~cm / 2.52$^{\circ}$) and the random augmentation baseline MaRepo$_s$ (3.2~cm / 1.54$^{\circ}$). Specifically, PoseCompass yields a 50.4\% relative improvement over the vanilla MaRepo baseline and a 43.0\% improvement over MaRepo$_s$. These consistent gains demonstrate that our value-based selection strategy effectively identifies high-quality, information-rich poses that target model weaknesses, offering superior supervision compared to random sampling.

\begin{table}[htbp]
\caption{Pose Estimation Results on Cambridge Landmarks. Median Translation/Rotation Error (cm/°). Best in \textbf{Bold}, Second Best \underline{Underlined}.}
\label{tab:cambridge}
\begin{center}
\footnotesize
\setlength{\tabcolsep}{2pt}
\begin{tabular}{l|cccc|c}
\toprule
\textbf{Methods} & \textbf{College} & \textbf{Hospital} & \textbf{Shop} & \textbf{Church} & \textbf{Avg.} \\
\midrule
PoseNet~\cite{kendall2015posenet} & 166/4.86 & 262/4.90 & 141/7.18 & 245/7.95 & 204/6.23 \\
MapNet~\cite{brahmbhatt2018geometry} & 107/1.89 & 194/3.91 & 149/4.22 & 200/4.53 & 163/3.64 \\
MS-Trans.~\cite{shavit2021learning} & 83/1.47 & 181/2.39 & 86/3.07 & 162/3.99 & 128/2.73 \\
PAE~\cite{shavit2022camera} & 90/1.49 & 207/2.58 & 99/3.88 & 164/4.16 & 140/3.03 \\
DFNet~\cite{chen2022dfnet} & 73/2.37 & 200/2.98 & 67/2.21 & 137/4.03 & 119/2.90 \\
PMNet~\cite{lin2024learning} & 68/1.97 & 103/1.31 & 58/2.10 & 133/3.73 & 90/2.27 \\
RAP~\cite{Li2025unleashing} & \underline{52/0.90} & \underline{87/1.21} & \underline{35/1.64} & \underline{53/1.52} & \underline{57/1.32} \\
MaRepo~\cite{chen2024mapose} & 58.24/1.01 & 97.44/1.36 & 39.20/1.84 & 59.36/1.70 & 63.56/1.48 \\
MaRepo$_s$~\cite{chen2024mapose} & 53.82/0.96 & 87.55/1.29 & 36.40/1.75 & 54.91/1.63 & 58.17/1.41 \\
\midrule
\textbf{PoseCompass} & \textbf{32.03/0.53} & \textbf{53.59/0.72} & \textbf{21.56/0.97} & \textbf{32.65/0.90} & \textbf{34.96/0.78} \\
\bottomrule
\end{tabular}
\end{center}
\end{table}

\textbf{Cambridge Landmarks~\cite{kendall2015posenet}.} 
On the outdoor dataset (Table~\ref{tab:cambridge}), PoseCompass reduces average error to 34.96 cm / 0.78°. The robust gains across diverse outdoor conditions validate that value-based selection generalizes beyond indoor scenes by balancing localization difficulty, coverage novelty, and rendering observability.

\begin{table}[t]
\centering
\small
\caption{\textbf{Ablation study on PoseCompass components.} We evaluate each component's contribution on the Chess scene from 7-Scenes. Results show median translation/rotation error (cm/°). All variants use $K=4000$ synthetic poses with MaRepo backbone.}
\label{tab:ablation_components}
\setlength{\tabcolsep}{3.5pt}
\begin{tabular}{l|ccc|c|cc}
\toprule
\textbf{Method} & $\mathcal{F}_{\text{diff}}$ & $\mathcal{F}_{\text{nov}}$ & $\mathcal{F}_{\text{gs}}$ & \textbf{Difix3D+} & \textbf{t (cm)} & \textbf{r (°)} \\
\midrule
Random sampling & - & - & - & - & 2.1 & 1.24 \\
\midrule
Only $\mathcal{F}_{\text{diff}}$ & \checkmark & - & - & - & 1.8 & 1.05 \\
Only $\mathcal{F}_{\text{nov}}$ & - & \checkmark & - & - & 2.0 & 1.18 \\
Only $\mathcal{F}_{\text{gs}}$ & - & - & \checkmark & - & 1.9 & 1.12 \\
\midrule
$\mathcal{F}_{\text{diff}}$ + $\mathcal{F}_{\text{nov}}$ & \checkmark & \checkmark & - & - & 1.6 & 0.95 \\
$\mathcal{F}_{\text{diff}}$ + $\mathcal{F}_{\text{gs}}$ & \checkmark & - & \checkmark & - & 1.4 & 0.82 \\
$\mathcal{F}_{\text{nov}}$ + $\mathcal{F}_{\text{gs}}$ & - & \checkmark & \checkmark & - & 1.7 & 0.98 \\
\midrule
All three components & \checkmark & \checkmark & \checkmark & - & 1.3 & 0.72 \\
\textbf{PoseCompass} & \checkmark & \checkmark & \checkmark & \checkmark & \textbf{1.2} & \textbf{0.65} \\
\bottomrule
\end{tabular}
\end{table}

\subsection{Ablation Study}

Table~\ref{tab:ablation_components} dissects PoseCompass on the Chess scene with $K{=}4000$ (equal to training set size). Random sampling yields 2.1 cm / 1.24°. Using only $\mathcal{F}_{\text{diff}}$ reduces error to 1.8 / 1.05, showing difficulty-driven selection targets model weaknesses. Dual-component variants achieve further gains: $\mathcal{F}_{\text{diff}}{+}\mathcal{F}_{\text{nov}}$ (1.6 / 0.95) balances hard examples with spatial diversity; $\mathcal{F}_{\text{diff}}{+}\mathcal{F}_{\text{gs}}$ (1.4 / 0.82) performs best by filtering low-quality renderings while prioritizing challenging regions. 

Combining all three dimensions achieves 1.3 / 0.72 (38.1\% / 41.9\% improvement over random), demonstrating complementarity across difficulty, novelty, and quality. Adding Difix3D+ appearance alignment further reduces error to 1.2 / 0.65, confirming that bridging the syn-to-real gap enhances supervision quality without altering pose selection.

We further analyze the robustness of PoseCompass to the weighting exponents $(\alpha, \beta, \gamma)$ in Eq.~(2). Varying these values within a broad range around the default setting leads to consistent pose rankings and similar localization accuracy, with performance variations below 5\% on Chess. This suggests that the exponents primarily encode relative preferences among difficulty, novelty, and observability, rather than requiring precise tuning.

\begin{table}[t]
\centering
\small
\caption{\textbf{Impact of real data ratio and synthetic budget on Chess scene.} We report median translation/rotation error (cm/°) and training time. $N{=}4000$ denotes the full training set size, $K$ is the synthetic pose budget. Results validate that both real data quality (for 3DGS reconstruction) and intelligent synthetic selection are essential.}
\label{tab:budget_analysis}
\setlength{\tabcolsep}{2.5pt}
\begin{tabular}{l|c|c|c}
\toprule
\textbf{Configuration} & \textbf{Error (cm/°)} & \textbf{Time (min)} & \textbf{Speedup} \\
\midrule
100\% real, no aug & 2.6/1.35 & 3.8 & - \\
\midrule
\multicolumn{4}{c}{\textit{Random Augmentation (baseline)}} \\
20\% real + random ($K{=}N$) & 2.8/1.52 & 12.5 & - \\
50\% real + random ($K{=}N$) & 2.4/1.38 & 13.8 & - \\
100\% real + random ($K{=}N$) & 2.1/1.24 & 15.2 & $1.0\times$ \\
\midrule
\multicolumn{4}{c}{\textit{PoseCompass: varying real data ratio}} \\
20\% real + ($K{=}N$) & 2.3/1.28 & 4.2 & $3.6\times$ \\
50\% real + ($K{=}N$) & 1.7/0.95 & 4.7 & $3.2\times$ \\
100\% real + ($K{=}N$) & \textbf{1.2/0.65} & 5.1 & $3.0\times$ \\
\midrule
\multicolumn{4}{c}{\textit{PoseCompass: varying synthetic budget}} \\
100\% real + ($K{=}0.2N$) & 1.8/1.02 & 4.1 & $3.7\times$ \\
100\% real + ($K{=}0.5N$) & 1.5/0.82 & 4.6 & $3.3\times$ \\
100\% real + ($K{=}N$) & \textbf{1.2/0.65} & 5.1 & $3.0\times$ \\
100\% real + ($K{=}2N$) & 1.2/0.64 & 9.8 & $1.6\times$ \\
\bottomrule
\end{tabular}
\end{table}

\subsection{Discussion on Data Synthesis}

\textbf{Rendering Budget and Real Data Efficiency.} 
Table~\ref{tab:budget_analysis} examines the interplay between real data ratio and synthetic budget on Chess. Two key findings emerge. First, while both methods benefit from more real data, PoseCompass exhibits steeper improvement curves than random augmentation. This reveals that intelligent selection better exploits high-quality 3DGS reconstruction—our observability filtering more effectively distinguishes artifact-free regions when the renderer is well-trained, and difficulty scoring more accurately identifies challenging viewpoints.

Second, value-based selection consistently outperforms random augmentation at matched real data ratios, with amplified advantages under limited data. Under sparse conditions, 3DGS produces uneven reconstruction quality. Random sampling blindly includes degraded regions, introducing noisy gradients. Our $\mathcal{F}_{\text{gs}}$ actively filters these artifacts while $\mathcal{F}_{\text{diff}}$ ensures coverage of challenging areas. The synthetic budget ablation shows diminishing returns beyond $K{=}N$, confirming efficient convergence. Consistent speedup stems from pose-level selection reducing redundant rendering while maintaining supervision quality.

\begin{table}[t]
\centering
\small
\caption{\textbf{Generalization across different APR backbones on Chess scene.} We apply PoseCompass ($K{=}N$) to multiple pose regression architectures. Results show median translation/rotation error (cm/°).}
\label{tab:backbone_comparison}
\setlength{\tabcolsep}{6pt}
\begin{tabular}{l|cc|cc}
\toprule
\multirow{2}{*}{\textbf{Backbone}} & \multicolumn{2}{c|}{\textbf{Baseline}} & \multicolumn{2}{c}{\textbf{+ PoseCompass}} \\
\cline{2-5} 
 & \textbf{t (cm)} & \textbf{r (°)} & \textbf{t (cm)} & \textbf{r (°)} \\
\midrule
PoseNet~\cite{kendall2015posenet} & 32 & 8.12 & 20.8 & 5.3 \\
MS-Transformer~\cite{shavit2021learning} & 11 & 4.66 & 7.2 & 3.0 \\
DFNet~\cite{chen2022dfnet} & 5 & 1.88 & 3.3 & 1.2 \\
MaRepo~\cite{chen2024mapose} & 2.6 & 1.35 & \textbf{1.2} & \textbf{0.65} \\
\bottomrule
\end{tabular}
\end{table}

\textbf{Backbone Generalizability.} 
Table~\ref{tab:backbone_comparison} examines PoseCompass across four APR architectures with varying representation capacities. All backbones benefit from value-based selection. However, the improvement magnitude exhibits a clear stratification. Lightweight models (PoseNet, MS-Transformer, DFNet) achieve consistent gains around 35\%. In contrast, MaRepo demonstrates 53.8\% improvement. 

This performance gap reveals a fundamental mechanism. Stronger pre-trained representations provide richer semantic features for localization. When combined with PoseCompass, such representations can better distinguish and exploit the complementary signals encoded in difficulty-driven and quality-filtered synthetic samples. Specifically, challenging viewpoints expose model blind spots through higher gradient magnitudes. Artifact-free renderings ensure these gradients remain informative rather than noisy. MaRepo's global attention mechanism amplifies this synergy by capturing long-range spatial dependencies that align with our value function's multi-dimensional scoring. The consistent plug-and-play transferability confirms that PoseCompass operates as a data-centric module independent of architectural inductive biases.

\section{Conclusion}
\label{sec:conclusion}




We propose PoseCompass, an intelligent synthetic pose selection pipeline for data-efficient APR training via 3D Gaussian Splatting. Our value function integrates Localization Difficulty to prioritize challenging regions, Coverage Novelty to ensure viewpoint diversity, and Rendering Observability to filter low-quality renderings, enabling efficient identification of high-value synthetic training poses while maintaining computational efficiency. Extensive experiments on 7-Scenes and Cambridge Landmarks demonstrate that PoseCompass significantly reduces median pose errors while achieving substantial training speedup over random augmentation. The pipeline is compatible with diverse renderers and backbones, demonstrating strong generalization across APR architectures.

\bibliographystyle{IEEEbib}
\bibliography{icme2025references}

\vspace{12pt}
\end{document}